\documentclass[conference]{IEEEtran}
\IEEEoverridecommandlockouts

\usepackage{tikz}

\newcommand\copyrighttext{%
  \footnotesize \textcopyright 2025 IEEE. Personal use of this material is permitted.
  Permission from IEEE must be obtained for all other uses, in any current or future
  media, including reprinting/republishing this material for advertising or promotional
  purposes, creating new collective works, for resale or redistribution to servers or
  lists, or reuse of any copyrighted component of this work in other works.}
\newcommand\copyrightnotice{%
\begin{tikzpicture}[remember picture,overlay]
\node[anchor=south,yshift=10pt] at (current page.south) 
  {\fbox{\parbox{\dimexpr\textwidth-\fboxsep-\fboxrule\relax}{\copyrighttext}}};
\end{tikzpicture}%
}

\usepackage{cite}
\usepackage{amsmath,amssymb,amsfonts}
\usepackage{algorithmic}
\usepackage{graphicx}
\usepackage{textcomp}
\usepackage{xcolor}
\def\BibTeX{{\rm B\kern-.05em{\sc i\kern-.025em b}\kern-.08em
    T\kern-.1667em\lower.7ex\hbox{E}\kern-.125emX}}
\begin{document}

\title{A Re-Calibration Method for Object Detection with Multimodal Alignment Bias in Autonomous Driving
}

\author{\IEEEauthorblockN{1\textsuperscript{st} Zhihang Song}
\IEEEauthorblockA{\textit{Department of Automation} \\
\textit{Tsinghua University}\\
Beijing \\
song-zh22@mails.tsinghua.edu.cn}
\\
\IEEEauthorblockN{4\textsuperscript{th} Lihui Peng}
\IEEEauthorblockA{\textit{Department of Automation} \\
\textit{Tsinghua University}\\
Beijing \\
lihuipeng@tsinghua.edu.cn}

\and
\IEEEauthorblockN{2\textsuperscript{nd} Dingyi Yao}
\IEEEauthorblockA{\textit{Department of Automation} \\
\textit{Tsinghua University}\\
Beijing \\
ydy24@mails.tsinghua.edu.cn}
\\
\IEEEauthorblockN{5\textsuperscript{th} Danya Yao}
\IEEEauthorblockA{\textit{Department of Automation} \\
\textit{Tsinghua University}\\
Beijing \\
 yaody@tsinghua.edu.cn}
 
\and
\IEEEauthorblockN{3\textsuperscript{rd} Ruibo Ming}
\IEEEauthorblockA{\textit{Department of Automation} \\
\textit{Tsinghua University}\\
Beijing \\
mrb22@mails.tsinghua.edu.cn}
\\
\IEEEauthorblockN{6\textsuperscript{th} Yi Zhang}
\IEEEauthorblockA{\textit{Department of Automation} \\
\textit{Tsinghua University}\\
Beijing \\
zhyi@tsinghua.edu.cn}

}

\maketitle
\copyrightnotice
\begin{abstract}
Multi-modal object detection in autonomous driving has achieved great breakthroughs due to the usage of fusing complementary information from different sensors. The calibration in fusion between sensors such as LiDAR and camera was always supposed to be precise in previous work. However, in reality, calibration matrices are fixed when the vehicles leave the factory, but mechanical vibration, road bumps, and data lags may cause calibration bias. As there is relatively limited research on the impact of calibration on fusion detection performance, multi-sensor detection methods with flexible calibration dependency have remained a key objective. In this paper, we systematically evaluate the sensitivity of the SOTA EPNet++ detection framework and prove that even slight bias on calibration can reduce the performance seriously. To address this vulnerability, we propose a re-calibration model to re-calibrate the misalignment in detection tasks. This model integrates LiDAR point cloud, camera image, and initial calibration matrix as inputs, generating re-calibrated bias through semantic segmentation guidance and a tailored loss function design. The re-calibration model can operate with existing detection algorithms, enhancing both robustness against calibration bias and overall object detection performance. Our approach establishes a foundational methodology for maintaining reliability in multi-modal perception systems under real-world calibration uncertainties.
\end{abstract}


\section{Introduction}
The object detection is a major element in the environment perception of autonomous driving. Many brilliant works have been made based on different sensor inputs such as camera\cite{ku2019monocular,liu2021autoshape,dong2025novel}, LiDAR\cite{shi2019pointrcnn,lang2019pointpillars,yin2021center}, or both of them\cite{xu2018pointfusion, wang2021pointaugmenting, liu2023bevfusion}. To better assure the safety and trustworthiness performance of autonomous driving perception, researchers have taken uncertainty quantification into consideration and one possible way is to use the multi-modal fusion method\cite{Quantification}. Due to the usage of complementary data from different sensors, such as image and point cloud, the multi-modal fusion method can achieve better performance than single-sensor ones in complex environments\cite{liu2023bevfusion}. However, besides the advantages, multi-modal fusion also brings more new challenges for algorithms. As the data representation from different sensors can differ significantly, a unique model structure design is a must to integrate features from each modal. Also, sensors in different locations must be calibrated to share the same world coordinate to correctly align and fuse information. Thus, accurate calibration is needed to reduce sensor uncertainty in perception results\cite{Quantification}. 

Previous works have made lots of breakthroughs for the first challenge by using sequential fusion\cite{zhao20193d}, bird's eye fusion\cite{liu2023bevfusion},\cite{cai2023bevfusion4d}, LiDAR-guided image fusion\cite{huang2020epnet}, etc. As for the sensor calibration problem, the calibration matrices are always taken as fixed and perfectly given in the 3D object detection process. Training for 3D object detection often uses well-processed datasets such as KITTI dataset\cite{geiger2013vision}, giving precise calibration matrices per frame. Some data collection vehicles such as in ONCE dataset\cite{mao2021one} are corrected daily for calibration, which is not the situation in real driving. Previous studies in calibration mainly focus on the calibration process before vehicles get out of the factory and the calibration is then fixed and doesn't need to be considered in perception tasks. However, in reality, vehicles on roads may encounter bumps and vibrations\cite{dong2023benchmarking}. Because the LiDAR scans its surroundings in a period but the camera takes images instantaneously, it might cause bias in calibration information too. There may also exist data delays and sensor drifts which can also influence the calibration performance\cite{dong2023benchmarking}.

Recently, some researchers have also paid attention to the calibration influence on 3D object detection performance. Two benchmarking works\cite{yu2023benchmarking, dong2023benchmarking}in 2023 indicated that alignment bias like other data corruptions such as weather and motion compensation, are challenging to fusion models. Since multi-modal fusion detection heavily depends on the feature alignment, the accuracy of calibration is of great importance. Yet, most calibration studies are designed for offline experiments with defined targets to provide the user with one-time precise calibration parameters, which need hours and specific experimental fields. Some recent targetless deep-learning-based calibration algorithms use iterative methods\cite{liu2021semalign} or Brute-force search\cite{luo2023calib} to complete the calibration correctness, but these methods also take several seconds or minutes to finish and can't be used in detection tasks. As the studies for calibration correctness in 3D object detection are relatively few, we developed a flexible and fast re-calibration detection framework to deal with calibration bias and noises.

Our work uses semantic segmentation features to guide the re-calibration inference and design a unique loss and training process. The re-calibration module takes the image, point cloud, and initial rough calibration as inputs and outputs corrected calibration before sending them into the 3D object detection module. In our experiments, we implemented the EPNet++\cite{liu2022epnet++} as the detection test model, which has an advanced performance in multi-modal algorithms. We conducted two types of calibration corruption experiments. One is the calibration matrices have random noise or bias, and the other is the LiDAR point cloud location in world coordinates has a small translation. We conducted a series of experiments and demonstrated that our method can significantly improve the performance of 3D detection method in both corruptions. The latest related work GraphAlign++\cite{graphalign} achieves great robustness in calibration of Gaussian noise by designing the perception algorithm structure. Our work came up with a modular re-calibration framework in another aspect, so it can be easily shifted to different detection methods as well as other calibration-dependant tasks without modifying perception models.

The main contributions of this work include: (a) Experiments in our study support the view that spatial misalignment has a great impact on 3D detection tasks. (b) We propose a simple but effective framework to re-calibrate the misalignment in detection tasks. (c) We design a fast and flexible re-calibration model and training method for the framework.


\section{Related Works} \label{Relatedworks}
\subsection{LiDAR-Camera 3D Object Detection}


Related works can be categorized mainly into three fusion types: point-level, feature-level, and proposal-level fusion. The point-level fusion method fuses the feature information obtained from the camera into the LiDAR point cloud and enhances the expression ability of points by projection. Early typical methods contain PointPainting\cite{vora2020pointpainting}, FusionPainting\cite{xu2021fusionpainting}, etc. These methods extract semantic information from the camera to augment each point and then use the point cloud detection head to complete the task. Other approaches such as EPNet\cite{huang2020epnet}, EPNet++\cite{liu2022epnet++}, PointAugmenting\cite{wang2021pointaugmenting}, and FocalsConv\cite{chen2022focal} implement richer features and more refined fusion methods to achieve better detection results. The feature-level fusion method extracts features from two input modalities and fuses them into one feature space by designed networks. Many outstanding studies implement Transformer-based structures to align features into LiDAR coordinates\cite{goodnough1999transfusion,li2022deepfusion,chen2023futr3d} or bird's eye view space\cite{liu2023bevfusion,cai2023bevfusion4d,graphalign,song2024graphbev}. The proposal-level fusion method typically involves two independent extractors to generate 3D bounding box predictions for both modalities and then use post-processing techniques to fuse them\cite{chen2017mv3d,qi2018frustum,pang2020clocs}. 

\subsection{LiDAR-Camera Calibration}

The calibration of extrinsic parameters is to determine the relationship between the positions and directions of different sensors. 


With the stunning development of deep learning, learning-based extrinsic calibration methods have emerged to solve the calibration problem. By using perception methods to extract features, end-to-end networks are developed to estimate calibration parameters from raw sensor inputs\cite{schneider2017regnet,iyer2018calibnet,yuan2020rggnet}. The odometry information is adopted in spatial and temporal parameter estimation\cite{taylor2016motion,park2020spatiotemporal}. In 2021, LCCNet was developed to predict the extrinsic parameters in real-time\cite{lv2021lccnet}. \cite{pervsic2021spatiotemporal} proposed a calibration method based on the Gaussian process estimated moving target trajectories. Such methods are limited by the odometry and tracking results\cite{luo2023calib}. Besides, semantic segmentation features are also widely adopted for calibrating LiDAR and cameras\cite{liu2021semalign,tsaregorodtsev2022extrinsic,rotter2022automatic}. Calib-Anything\cite{kirillov2023segment} 
and utilizes the large model SAM to deal with the input data and uses the Brute-force searching method for calibration\cite{luo2023calib}. Some methods such as SST-Calib also used combined features of visual odometry and semantic segmentation to infer spatial-temporal parameters\cite{kodaira2022sst}. In 2024, CalibFormer implemented a transformer-based correlation head to align features for calibration\cite{xiao2023calibformer}. 

\section{Methodology} \label{Methodology}



\subsection{Data Preparation} \label{dataprepare}

In order to train the re-calibration module, we first need to establish training datasets with calibration bias inputs and precise labels. We choose the KITTI dataset\cite{geiger2013vision} as our data source and design two types of alignment bias to augment our training set, i.e. random noise and point cloud translation.

\textbf{Random Noise:} The first type is random noise in calibration matrices. This kind of noise can represent the bias in calibration initialization and the latter changes in driving bumps and vibration. We make this kind of data by adding different Gaussian noise matrices with variance $\sigma^2$ to the extrinsic matrix. However, this kind of noise is not enough for re-calibration training because all the calibration labels in KITTI are close to several specific matrices which may lead to a short-cut learning of regression to fixed values. Also, the Gaussian noise can not represent to point cloud location bias caused by the stuck or drifting sensor. Thus, we also conduct the point cloud translation as the second data generation method.

\textbf{Point Cloud Translation:} As for the point cloud translation, we use the translation matrices and their reverse matrices to generate new point clouds and corresponding labels. The generation is shown as (\ref{eq-data}) below.

\begin{figure}[hb] 
 \centering
 \small
 \setlength{\arraycolsep}{2pt}
    \begin{equation}
    {
    \left[ \begin{array}{c}
    x' \\
    y' \\
    z' \\
    1
    \end{array} 
    \right ]}^{T}
    =
    \underbrace{
    {\left[ \begin{array}{c}
    x \\
    y \\
    z \\
    1
    \end{array} 
    \right ]}^{T}
    *
    \mathop{{
    \left[ \begin{array}{cccc}
    1 & 0 & 0 & 0\\
    0& 1 & 0 & 0\\
    0& 0 & 1 & 0\\
    a&b&c&1
    \end{array}
    \right ]}}\limits_{Tr}
    }_{New\ point\ cloud}
    *
    \underbrace{Tr^{-1}
    *V2C^{T}}_{New\ extrinsic\ label}
    *R0^{T}
    \label{eq-data}
    \end{equation}
    \setlength{\arraycolsep}{5pt}
\end{figure}

$a, b, c$ are random translation parameters, $V2C$ and $R0$ are the extrinsic and rectified matrix. As EPNet++ is based on the point cloud, we also need to translate the object labels simultaneously when combined in the detection evaluation process. 


\subsection{Re-calibration Framework} 

\begin{figure*}[htbp]
\centerline{\includegraphics[width=0.8\linewidth]{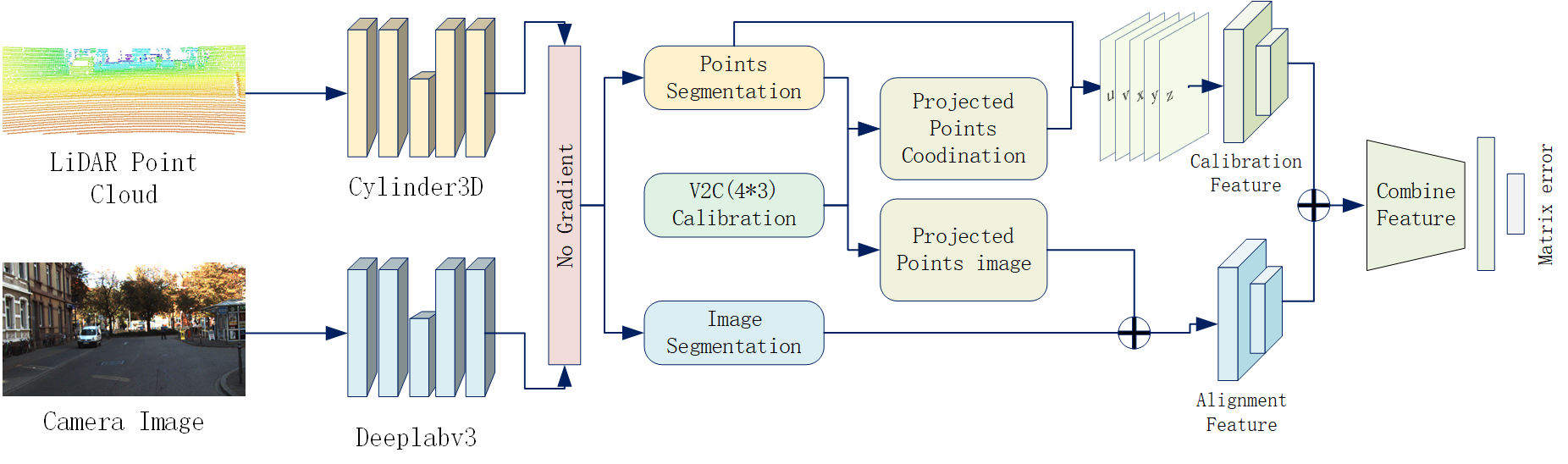}}
\caption{Framework of re-calibration model. Our model takes LiDAR point cloud, image and original calibration matrix as inputs and outputs re-calibrated bias.}
\label{fig: structure}
\end{figure*}

For the re-calibration model, we design the algorithm based on semantic-segmentation methods. The overall framework of our re-calibration model is shown in Figure \ref{fig: structure}.

In the first stage, the point cloud and image are sent into two pre-trained semantic segmentation models separately. In our experiments, we chose Cylinder3D\cite{pvkd,cylinder3d,cylinder3d-tpami} for point cloud segmentation and Deeplabv3\cite{chen2017rethinking} for image segmentation, which can be replaced by any needed segmentation methods. They are not involved in gradient calculation and parameter update. These segmentation models then output the interested segmentation labels which are determined by the detection tasks. In our experiments, we separate the segmentation of cars as inputs for the next stage.

In the second stage, the point cloud segmentation results are projected to the camera pixel coordinates by the input calibration matrices with bias. The projected point segmentation result is then concatenated with image segmentation to generate the alignment feature. The alignment feature contains segmentation information from points and images in the same pixel coordinate system. Thus the clusters of projected points may have displacements with corresponding image areas, which can guide the network to learn the bias of the projection.

Besides, we also want the model to learn the relationship between the input projection matrix parameters and the coordinates bias. Thus, at the same time, we also record the corresponding coordinates in 3D LiDAR coordinate system and 2D camera pixel coordinate system. Then, we select the interested points with the above corresponding coordinates to generate the calibration feature. The calibration feature is arranged in the pixel plane with five position values. $x,y,z$ are the original positions of LiDAR points, and $u,v$ are projected pixel positions. In this way, we involved the input projection matrix information in the learning process. The alignment feature and calibration feature is then fed into the fusion net which outputs the extrinsic matrix bias.

\subsection{Loss Design} 

\begin{figure}[htbp]
\centerline{\includegraphics[width=0.5\linewidth]{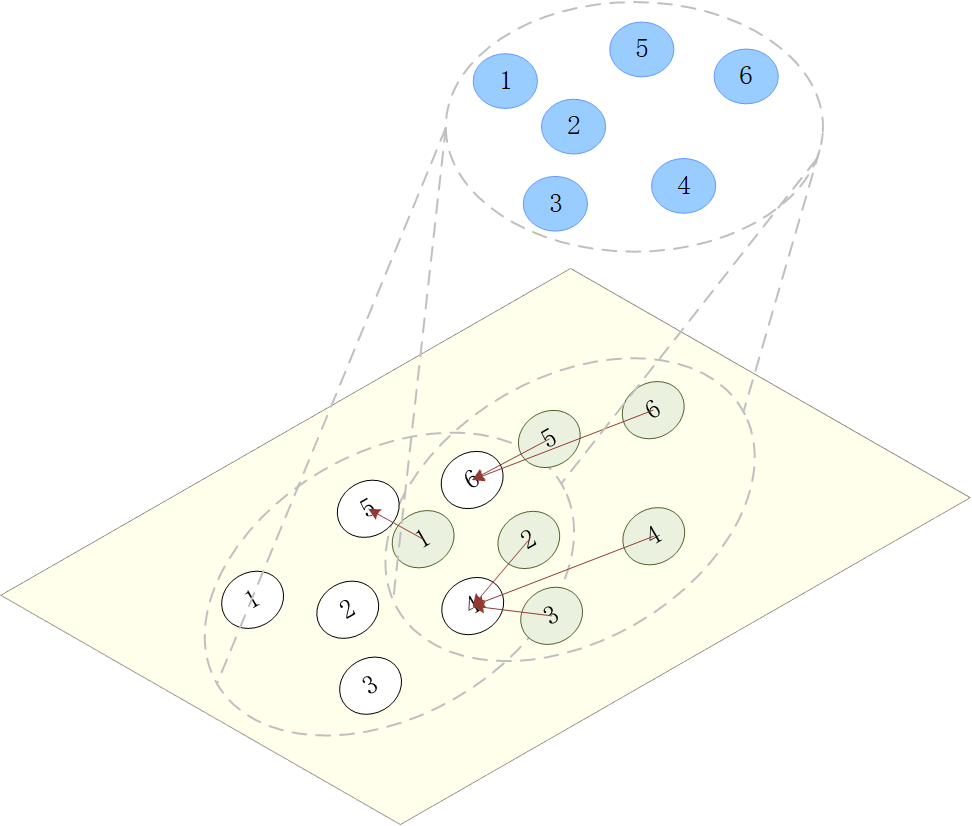}}
\caption{Projected loss. White points are the correct projected position by label calibration. Green points are projected position by output re-calibration results.}
\label{fig: loss}
\end{figure}

In re-calibration model training, we utilize a composite loss function with two parts. The first one is projected distance loss inspired by semantic alignment loss in SemAlign\cite{liu2021semalign}, which uses segmentation information to evaluate the quality of calibration projection. The semantic alignment loss is designed to calculate the distance between the projected point cloud and the same feature in the corresponding image. As our method is supervised and generates variance data with calibration label in \ref{dataprepare}, we can simplify this loss to calculate the distance between output and label projected point clusters. See Figure  \ref{fig: loss}. Projected loss calculates the distance between the points in clusters with the same interested class label. For each re-calibration projected point of a specific class, we calculate its distance to the nearest neighbor in the label projected points clusters. Then we use the sum value of distance as the projected loss. This kind of loss contributes to the model training in learning which elements in the matrix to modify and how they affect projection. Projected loss can be written in (\ref{eq1}). $p_{c}^{i}$ refers to the $i_{th}$ point in projected point cluster of re-calibration and $p_{l}^{j}$ refers to the $j_{th}$ point in label projected point cluster. $dist()$ calculates the L2 norm of displacement between two position vectors.

\begin{equation} \label{eq1}
    \mathcal{L}_\text{Pro}=\sum\limits_{i}{\mathop{min}\limits_{j}{(dist(p_{c}^{i},p_{l}^j))}}
\end{equation}

The other part of our loss function is a normal MSE loss between inference bias and label bias to each element in the extrinsic matrix. This part directly leads the model to learn where to optimize. As the projection from the LiDAR world coordinate system to 2D pixel coordinate system is a dimensionality reduction process, there exists infinite solutions to the projection alignment matrix. Thus, it's necessary to modify the weights of two losses because the projected loss may lead to another local optimum with a similar 2D projection but totally different calibration in 3D. The whole loss design can be expressed as (\ref{eq22}).
  \begin{equation} \label{eq22}
  \mathcal{L}_\text{Total} = \lambda_1  \text{MSE}(\text{calib}_{\text{out}} + \text{calib}_{\text{in}}, \text{calib}_{\text{label}})
 + \lambda_2 \mathcal{L}_\text{Pro}
  \end{equation}

In the experiment, as the projected loss is about $10^3$ to $10^4$ times of MSE loss, we set $\lambda_{1}$ and $\lambda_{2}$ to be 10 and 0.001 to make the model quickly converge. Then, we modify the $\lambda_{1}$ and $\lambda_{2}$ to be 10 and 0.00001 to make the MSE loss dominant and let the model converge to the calibration bias we need. If without the second stage, we found that though the re-calibration loss was minimized, the output re-calibrated matrix has little improvement on detection results (re-calibration result AP@0.7 bbox is 60.92 compared to 58.36 without re-calibration in Table \ref{tab2}).

\section{Experiment Results} \label{Experimentresults}
\begin{figure}[htbp]
\centerline{\includegraphics[width=0.7\linewidth]{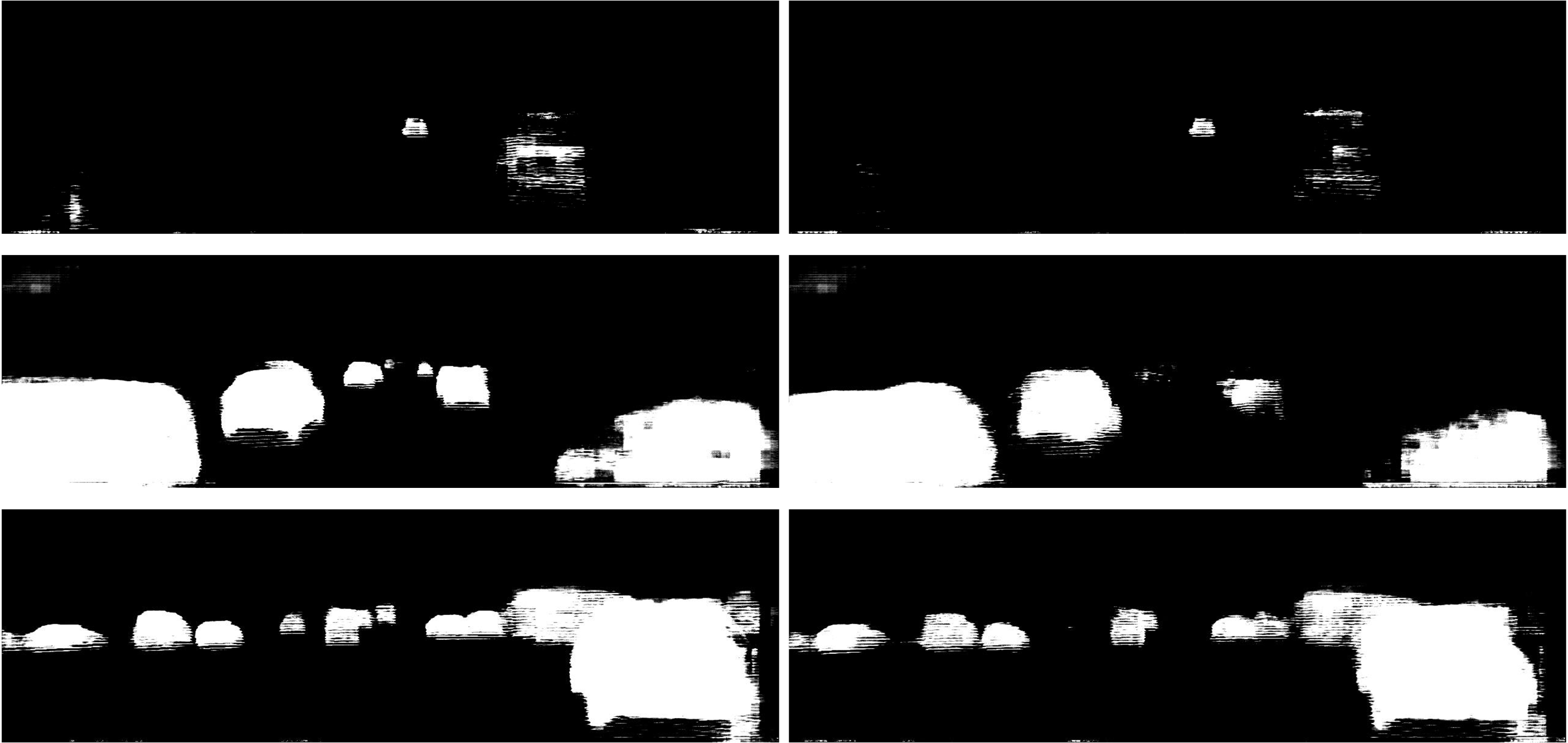}}
\caption{Corrupted feature fusion caused by Gaussian noise in calibration.(Left: without noise. Right: with noise.)}
\label{fig: gaussian}
\vspace{-1em}
\end{figure}

\begin{figure}[htbp]
\centerline{\includegraphics[width=0.7\linewidth]{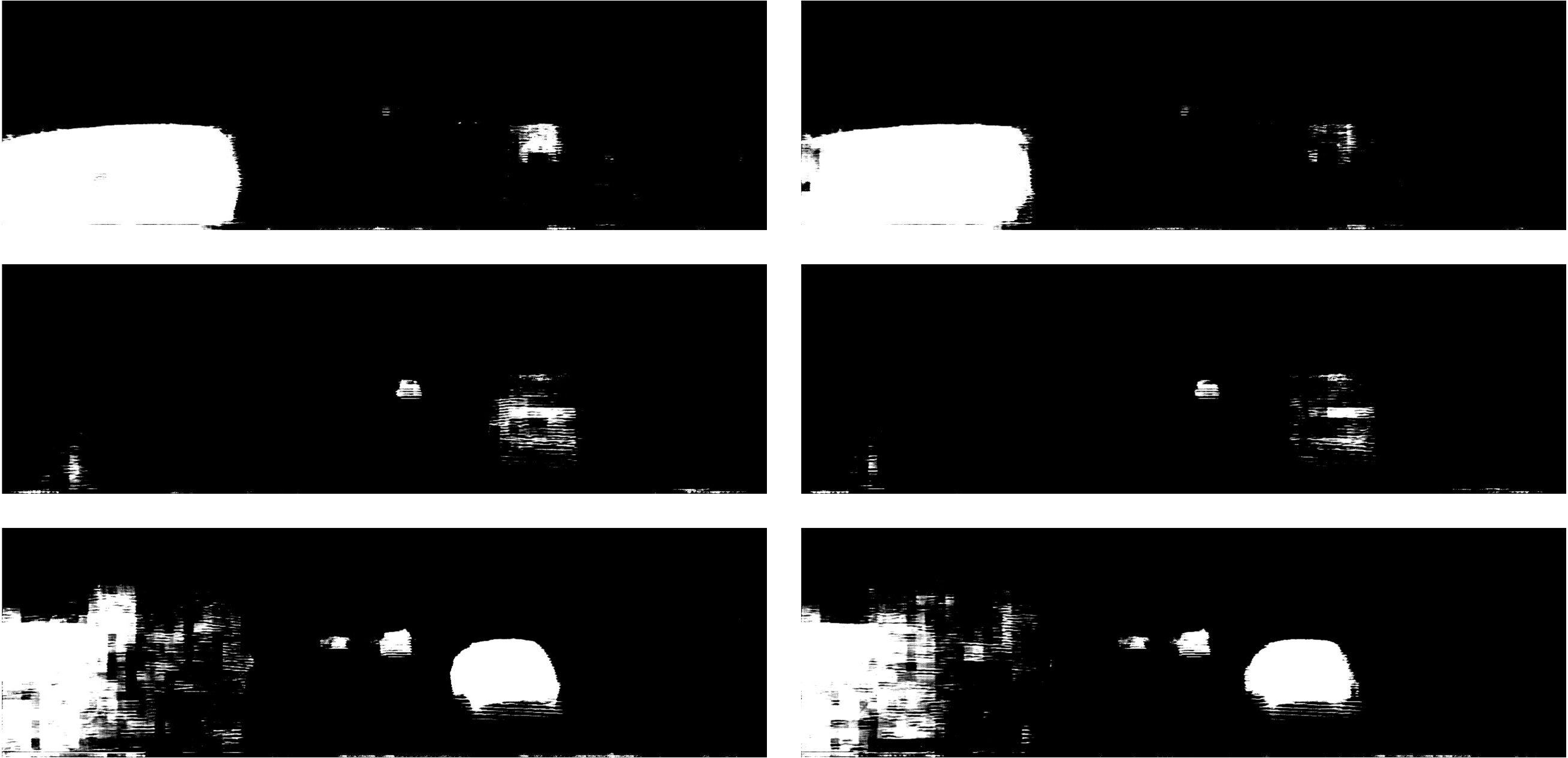}}
\caption{Corrupted feature fusion caused by point translation in LiDAR.(Left: without translation. Right: with translation.)}
\label{fig: lidar}
\vspace{-2em}
\end{figure}

To verify the re-calibration effects on multi-modal object detection methods, we conduct experiments with the EPNet++ algorithm. Firstly, we test the EPNet++ on the car category in KITTI dataset. By testing the same model with clean calibration, random Gaussian noise and point cloud translation, we proved that little bias in calibration may cause a severe performance drop for the multi-modal detection method using calibration matrices as direct inputs. The results are shown by giving a comparison in Table \ref{tab1}. With the Gaussian noise of $\sigma=0.01$ added to each element in original calibration matrices, all the four AP performances are corrupted. That's because the detection algorithm can't align the features from images and LiDAR points. See Figure \ref{fig: gaussian} and \ref{fig: lidar}, we extracted the fusion feature from RPN net in EPNet++, which shows the impact of calibration bias on the perception algorithm. With added noise, the features of vehicles got blurred or even vanished. The translation in LiDAR points also causes displacement and blur in the fusion feature. Thus, we infer that the severe misaligned fusion feature can't be identified by the fusion detection and causes low recall and AP.

\begin{table}[htbp]
\begin{center}
\caption{Comparison of EPNet++ performance (AP@0.7) with different calibration bias(Easy)}
\label{tab1}

\begin{tabular}{l*{5}{c}}
\hline

\textbf{\textit{Data Type}} & 
\textbf{\textit{bbox}}
&\textbf{\textit{bev}}& \textbf{\textit{3d}}& \textbf{\textit{aos}}&\\

\hline
Orig. data+calib. &99.21&96.13&92.79&99.05 \\
Orig. data+Gauss. noise calib. &58.36&52.03&18.08&58.28 \\
LiDAR $\delta$y=0.2m+trans. calib. &99.22&96.13&92.78&99.05 \\
LiDAR $\delta$y=0.2m+origi. calib. &95.37&72.18&42.75&95.19 \\
\hline
\end{tabular}
\end{center}
\end{table}

\begin{table}[htbp]
\begin{center}
\caption{Comparison of EPNet++ performance (AP@0.7) with different calibration bias(Moderate)}
\label{tab1-2}

\begin{tabular}{l*{5}{c}}
\hline

\textbf{\textit{Data Type}} & 
\textbf{\textit{bbox}}
&\textbf{\textit{bev}}& \textbf{\textit{3d}}& \textbf{\textit{ aos}}&\\

\hline
Orig. data+calib. &94.19&89.00&83.24&93.17 \\
Orig. data+Gauss. noise calib.  &47.30&42.02&15.19&47.15 \\
LiDAR $\delta$y=0.2m+trans. calib.  &94.19&89.00&83.24&93.17 \\
LiDAR $\delta$y=0.2m+origi. calib.  &89.05&66.68&39.20&88.56 \\
\hline
\end{tabular}
\end{center}
\end{table}

\begin{figure}[htbp]
\centerline{\includegraphics[width=0.7\linewidth]{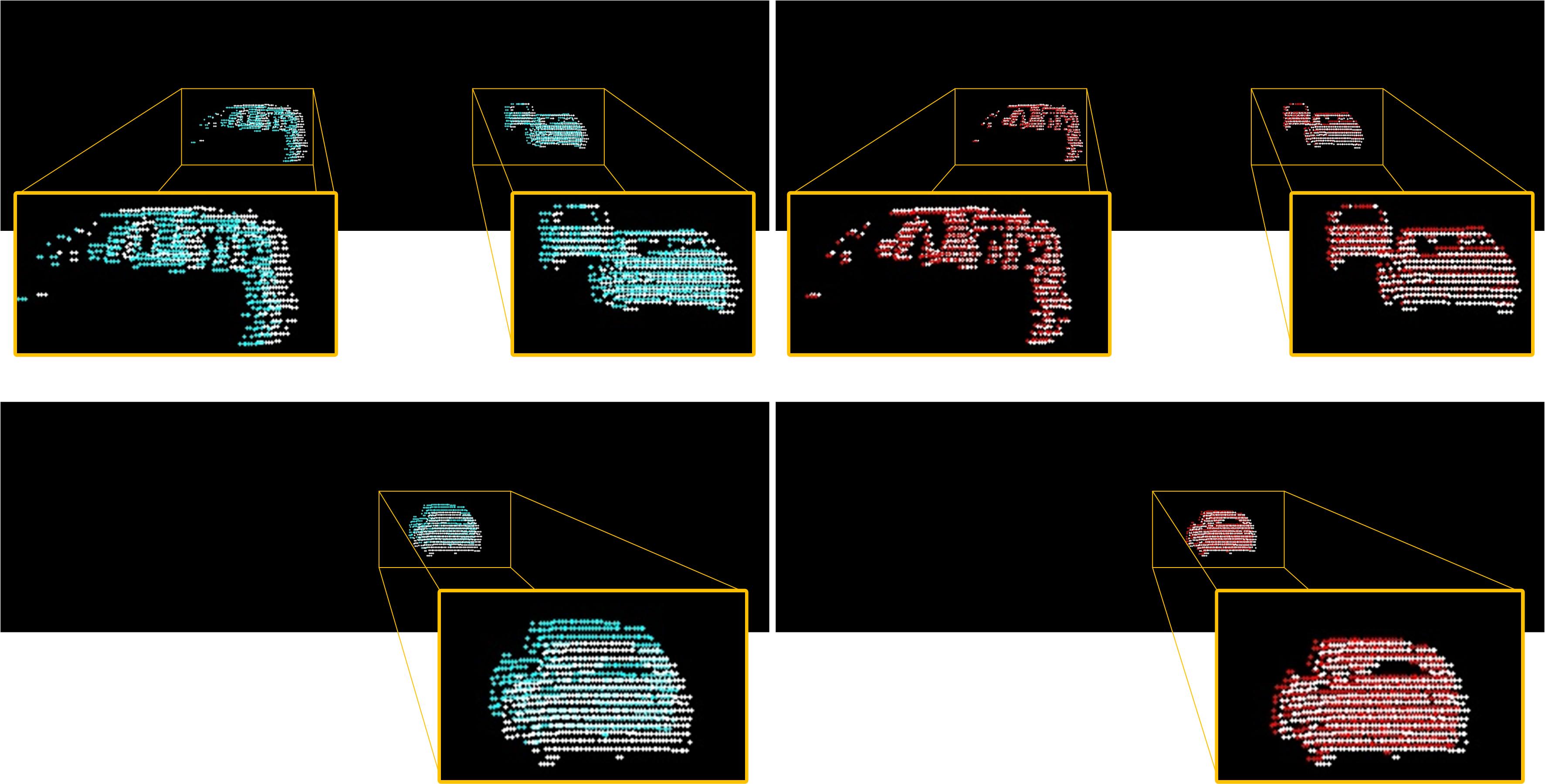}}
\caption{Comparison of points projection with Gaussian noise in calibration before and after re-calibration.(White: label; Green: before re-calibration; Blue: after re-calibration.)}
\label{fig: recalib-gaussian}
\end{figure}

\begin{figure}[htbp]

\centerline{\includegraphics[width=0.7\linewidth]{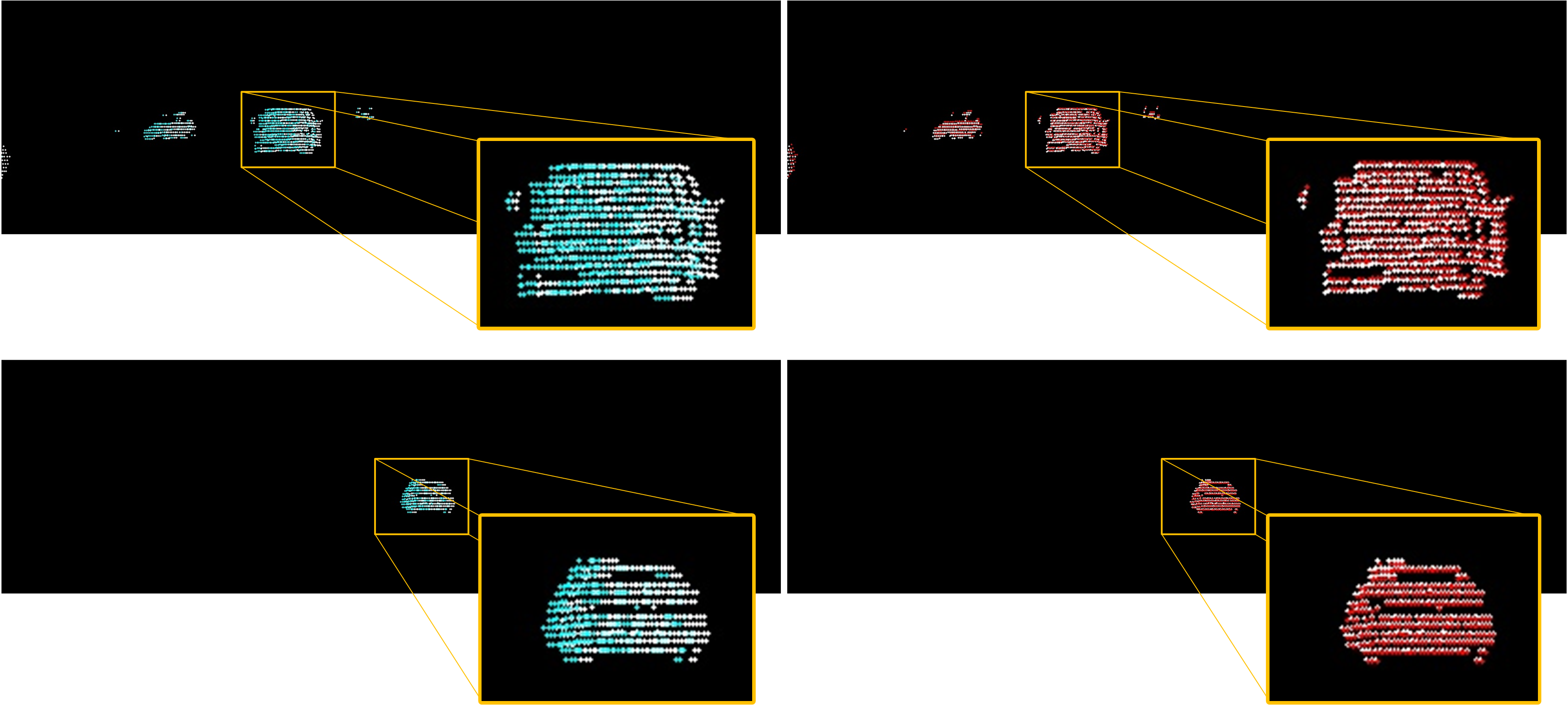}}
\caption{Comparison of points projection with point translation in LiDAR before and after re-calibration.(White: label; Green: before re-calibration; Blue: after re-calibration.)}
\label{fig: recalib-lidar}
\end{figure}

\begin{figure}[htbp]
\centerline{\includegraphics[width=0.7\linewidth]{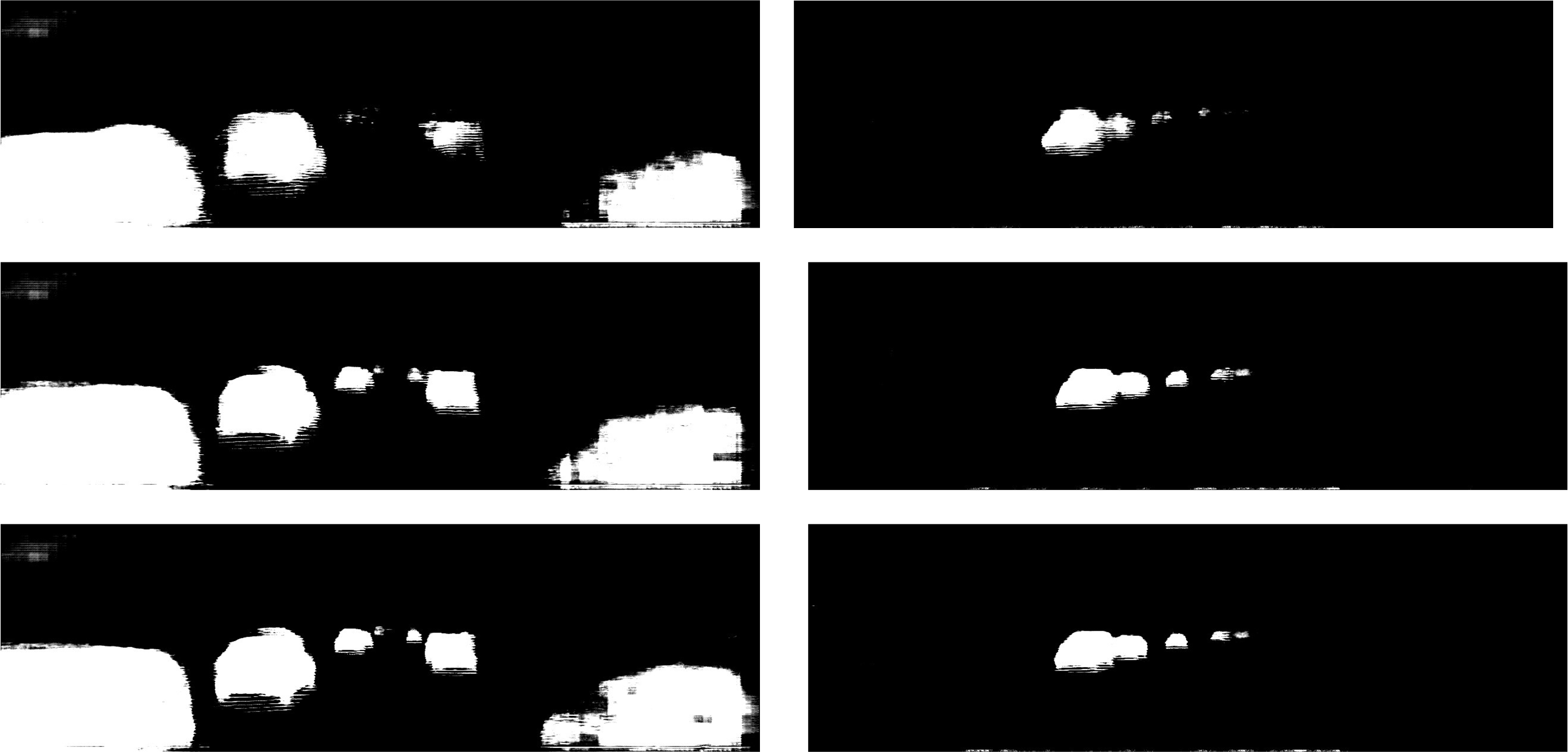}}
\caption{Comparison of RPN segmentation feature before and after re-calibration.(Up: before recalibration; Middle: after re-calibration; Down: ground truth.)}
\label{fig: realib-rpn}
\end{figure}

Then we applied our re-calibration module combined with the detection process and tested the performance again. The re-calibration result is shown in Table \ref{tab2}, Table  \ref{tab3}. From the table, we can see that the AP of the four metrics showed significant improvement in the Gaussian noise situation. The AP of 3D and BEV are improved in LiDAR translation situation. After the re-calibration process, although the detection method still can't recover to the ideal level, the performance is more acceptable for serious corruption. We can also see the re-calibration effect from the LiDAR point projected to image coordinates, see Figure \ref{fig: recalib-gaussian} and Figure \ref{fig: recalib-lidar}. It's obvious that the points we are interested in have less distance to the label, which makes the feature from the image and point cloud align better. We also visualize the RPN feature to confirm the recalibration effect on feature fusion, see Figure \ref{fig: realib-rpn}.

\begin{table}[htbp]

\begin{center}
\caption{Comparison of EPNet++ performance(AP@0.7) with re-calibration (Easy)}
\label{tab2}

\begin{tabular}{l*{5}{c}}
\hline

\textbf{\textit{Data Type}} & 
\textbf{\textit{bbox}}
&\textbf{\textit{bev}}& \textbf{\textit{3d}}& \textbf{\textit{aos}}&\\

\hline
Orig. data+Gauss. noise calib.
&58.36&52.03&18.08&58.28 \\
Orig. data+Gauss. noise re-calib. &83.48&57.67&35.38&83.37 \\

LiDAR $\delta$y=0.2m+origi. calib. &95.37&72.18&42.75&95.19 \\
LiDAR $\delta$y=0.2m+re-calib. &95.30&80.98&59.13&95.18 \\
\hline
\end{tabular}
\end{center}
\end{table}

\begin{table}[htbp]

\begin{center}
\caption{Comparison of EPNet++ performance(AP@0.7) with re-calibration (Moderate)}
\label{tab3}

\begin{tabular}{l*{5}{c}}
\hline

\textbf{\textit{Data Type}} & 
\textbf{\textit{ bbox}}
&\textbf{\textit{bev}}& \textbf{\textit{3d}}& \textbf{\textit{aos}}&\\

\hline
Orig. data+Gauss. noise calib.&47.30&42.02&15.19&47.15 \\
Orig. data+Gauss. noise re-calib.  &71.61&45.82&27.01&71.37 \\

LiDAR $\delta$y=0.2m+origi. calib. &89.05&66.68&39.20&88.56 \\
LiDAR $\delta$y=0.2m+re-calib.  &85.44&70.55&46.61&85.05 \\
\hline
\end{tabular}
\end{center}
\end{table}

In our experiment, we also found that the improvements in moderate objects and hard objects are not significant enough in comparison to those in easy ones. We analyzed that it might be because the semantic segmentation results are less accurate in the re-calibration model, which influences the feature extraction and alignment to calculate the calibration bias. Thus, one way to further improve the performance of re-calibration detection is to change a better pre-trained image segmentation algorithm and point cloud segmentation algorithm to produce precise segmentation information for calibration.

We also compared the performance and speed of our re-calibration model on the KITTI dataset. See the results with other segmentation-based calibration method in Table \ref{tab4}. As to the speed, except for the segmentation part without training, the main part of the re-calibration net takes about 40 ms on the NVIDIA RTX3090Ti GPU. The whole time of re-calibration work can be modified by changing the segmentation part and with the Deeplabv3 and Cylinder3D we used, it takes about 0.2-0.4 seconds per frame. Compared to 15-30 seconds consumed by segmentation methods such as SemAlign\cite{liu2021semalign} and CalibAnyThing\cite{luo2023calib}, our work has great speed advantages. 

\begin{table}[htbp]
\begin{center}
\caption{Comparison of segmentation re-calibration methods results}
\label{tab4}

\begin{tabular}{l*{3}{c}}
\hline

\textbf{\textit{Method}}
&\textbf{\textit{Translation Error(cm)}}& \textbf{\textit{Rotation Error($^{\circ}$)}}\\

\hline
\cite{tsaregorodtsev2022extrinsic} &20.2&0.34 \\
\cite{luo2023calib} &10.7&0.17 \\
Ours &10.3&0.21 \\

\hline
\end{tabular}
\end{center}
\end{table}

\section{Conclusion} \label{Conclusion}
This paper provides a flexible re-calibration detection framework to deal with the alignment bias problem in multi-modal perception tasks. Our approach utilizes semantic segmentation information to predict the extrinsic matrix bias and investigates how the calibration bias impacts the detection tasks. We implement our framework to EPNet++ and prove its usability. We hope that our work can provide a new perspective on improving the performance of existing multi-modal detection methods in misalignment bias. Our work focuses on spatial calibration. Future work could explore joint spatio-temporal re-calibration. Additionally, future work may integrate uncertainty-aware methods to boost model performance.


\bibliographystyle{IEEEtran}  
\bibliography{references}  

\end{document}